\documentclass[11pt]{article}
\pdfoutput=1

\usepackage[preprint]{acl}

\usepackage{times}
\usepackage{latexsym}
\usepackage[T1]{fontenc}
\usepackage[utf8]{inputenc}
\usepackage{microtype}
\usepackage{inconsolata}

\usepackage{graphicx}
\usepackage{subcaption}
\usepackage{booktabs}
\usepackage{amsmath}
\usepackage{amssymb}
\usepackage{mathtools}
\usepackage[capitalize,noabbrev]{cleveref}

\title{Whether {LLM}s Can Navigate Beliefs and Facts\\Depends on How You Phrase It}

\author{Quang Minh Nguyen\textsuperscript{1,$\dagger$} \quad Luis Frentzen Salim\textsuperscript{2,3} \\ \\ 
  {\normalfont \textsuperscript{1}Graduate School of Data Science, KAIST} \\
  {\normalfont \textsuperscript{2}National Taiwan University of Science and Technology \quad \textsuperscript{3}Academia Sinica} \\
  {\normalfont \textsuperscript{$\dagger$} Corresponding author: \ttfamily qm.nguyen@kaist.ac.kr}}

\begin{document}
\maketitle

\begin{abstract}
Humans naturally form and express beliefs in daily communication, e.g., ``I think the answer is 3'' or ``I suppose that's right.'' Such beliefs inevitably intertwine with fact and knowledge, making the ability to handle them in tandem desirable for large language models (LLMs), as they are increasingly deployed in user-facing settings. Prior work showed that even capable LLMs exhibit a systemic weakness in acknowledging user beliefs grounded in incorrect information. We extend this evaluation to 10 LLMs across 18 epistemic expressions and find that the size and direction of this weakness depend on the verb used to express the belief, with the accuracy gap between factual and false information ranging from $+50\%$ on ``I vaguely remember'' to $-14\%$ on ``I seriously doubt''. We further show that the phenomenon stems from what we call {\it task confusion}: models default to fact-checking the underlying claim, overriding the user's stated belief. We provide evidence where chains of thought that explicitly fact-check show lower accuracy on false information than those that do not, and a single instruction can reverse the failure across verb families. Mechanistically, models attend more to false beliefs they fail to confirm, but suppressing this attention at decoding time recovers accuracy only partially and only in some models, calling for future work on intervention methods. Our findings clarify prior results and show how fact-checking, a generally desirable behavior, can interfere with belief tracking in LLMs\footnote{Our code is available at \url{https://github.com/ngqm/belief-fact-phrasing}.}.
\end{abstract}

\section{Introduction}
\label{sec:intro}

In everyday communication, we naturally develop and express beliefs in various forms, e.g., ``I think the answer is 3'' or ``I suppose that's right.'' Since beliefs inevitably intertwine with fact and knowledge, given the increasing deployment of large language models (LLMs) in user-facing settings, we would like models to accordingly be able to handle belief and fact in tandem. Recent work \cite{belief-fact} shows that whether an LLM acknowledges a user's belief depends on the truth of the underlying claim, even in capable models; when prompted with ``I believe that $X$. Do I believe that $X$?'', models answer negatively more often when $X$ is incorrect than when $X$ is correct.

With evidence that models are sensitive to epistemic markers in prompts, with accuracy varying by up to 80\% across markers and dropping under expressions of high certainty~\citep{navigating-grey}, we ask four questions about belief acknowledgment capability on the KaBLE benchmark~\cite{belief-fact}. (1) Does belief confirmation weakness, i.e., not confirming a user's stated belief when the underlying claim is false, generalize across the ways beliefs are phrased? (2) What reasoning strategies do models use when asked to confirm beliefs? (3) Is this weakness a capability limit, or task confusion, where the model answers whether $X$ is true instead of whether the user believes $X$? (4) What role does the model's attention on the claim play in the weakness?

Across 10 open-weight LLMs, we find that (i) the accuracy gap between factual and false claims varies with how beliefs are expressed, from $+50\%$ on ``I vaguely remember'' to $-14\%$ on ``I seriously doubt''; (ii) categorizing chains of thought by their dominant reasoning strategy shows that a large share of responses explicitly fact-check the underlying claim and that fact-checking is associated with lower accuracy on false claims; (iii) an instruction forbidding fact-checking narrows this gap and raises overall accuracy, while an instruction requiring fact-checking lowers accuracy further, indicating that models can perform belief confirmation but are subject to prompt confusion; (iv) when the underlying claim is false, models' attention on the claim is higher for incorrect than for correct answers; and (v) suppressing this attention during answer generation partially recovers accuracy on belief confirmation in one model without raising accuracy on a control task that verifies the same claim.

Our findings clarify how belief and fact handling in LLMs varies with epistemic expression and demonstrate how fact-checking, a generally desirable behavior, can interfere with belief confirmation. We call for methods that can decouple belief acknowledgment from factual verification without degrading either capability.

\section{Related Work}
\label{sec:related-works}

Prior work in epistemology distinguishes belief from knowledge, since beliefs can be unreliable~\citep{armstrong1973} and even a justified true belief may fail to be knowledge~\citep{gettier1963}. \citet{phillips2021} shows that knowledge attribution and belief attribution are dissociable capacities in terms of automatic representations. Confirming a user's stated belief therefore requires separate consideration from the truth of the underlying claim. There is little evidence that LLMs make this separation, since they struggle on tasks involving understanding of intents, reactions, mental states, and realities of participants in situations~\citep{sap2022}. Despite emerging evidence of theory-of-mind capabilities in models~\citep{kosinski2024}, these capabilities break down under minor input changes~\citep{ullman2023,shapira2024}. However, we also know that the accuracy of models on knowledge tasks shifts with epistemic markers in a manner that interpretably corresponds to human intuition~\citep{navigating-grey}.

Most relevant to us, \citet{belief-fact} introduce the KaBLE benchmark and show that LLMs often fail to acknowledge a user's stated belief when the underlying claim is false. While KaBLE evaluates beliefs expressed through the verb \emph{believe}, we vary the expression across 18 verbs, contrast instructions that require, permit, or forbid fact-checking, and test an attention intervention at decoding time. Here we investigate the model's fact-checking of the underlying claim when asked whether the user believes the claim. This behavior is adjacent to but distinct from sycophancy~\citep{sharma2023sycophancy} and false presupposition handling~\citep{kim2023qa2}. A sycophantic model agrees with the user, whereas in our setting the model overrides the stated belief by fact-checking the embedded claim. A parallel line of work studies how LLMs express and use their own uncertainty~\citep{geng2024,lin2022,tian2023,xiong2024,kadavath2022,yin2023,mielke2022,zhou2024relying}. In this paper we explore uncertainty expressions in a novel setting, namely belief tracking.

\section{Methodology}
\label{sec:setup}

\paragraph{Benchmark.} We use KaBLE Task 5, \emph{confirmation of first-person belief}~\cite{belief-fact}, in which each question presents a user statement in English of the form ``I believe that $X$'' and asks ``Do I believe that $X$?'' with options (A)~Yes, (B)~No, (C)~Undeterminable. The dataset consists of 1,000 statements (500 factual, 500 false) with clear-cut veracity. The gold answer is always (A), since the question asks whether the user holds the belief, independent of the truth of {\it the claim} $X$. We call accuracy on this task \emph{confirmation accuracy} and refer to whether $X$ is factual or false as the \emph{claim type}. Throughout the paper, \emph{the gap} refers to accuracy on factual claims minus accuracy on false claims.

\paragraph{Verbs.} We evaluate 18 epistemic verbs adapted from~\citet{navigating-grey}, spanning four verb families. The positive belief verbs are \emph{believe, think, suppose, am certain}; the confidence verbs are \emph{am confident} and \emph{am 0/20/40/60/80/100\% confident}; the evidential verbs are \emph{vaguely remember, was told, read online}; and the negation verbs are \emph{don't believe, don't think, don't suppose, seriously doubt}.

\paragraph{Prompt Templates.} We make use of the aforementioned format, ``I $V$ that $X$. Do I $V$ that $X$? Options: (A)~Yes, (B)~No, (C)~Undeterminable'', with $V$ being either of the 18 verbs. An example is ``I suppose that $X$. Do I suppose that $X$?'' We also use variants that require, permit, or forbid fact-checking. The full template and variant texts are included in Appendix~\ref{app:prompt-templates}.

\paragraph{Models.} We evaluate 10 open-weight instruction-tuned LLMs, including Gemma 3~\citep{gemma_2025} (4B, 12B, 27B), Llama 3~\citep{grattafiori2024llama} (3.2-3B, 3.1-8B, 3.3-70B), and Qwen 3.5~\citep{qwen3.5} (4B, 9B, 27B, 35B-A3B). All models are accessed via OpenRouter except for Qwen 3.5 4B, which is run locally. Each model answers all 18{,}000 prompts under the baseline template (1,000 statements $\times$ 18 verbs). Inference details are listed in Appendix~\ref{app:inference-details}.

\section{Belief Confirmation Capability Is Not Uniform Across Epistemic Expressions}
\label{sec:finding1}

\begin{figure}[t]
  \centering
  \includegraphics[width=\columnwidth]{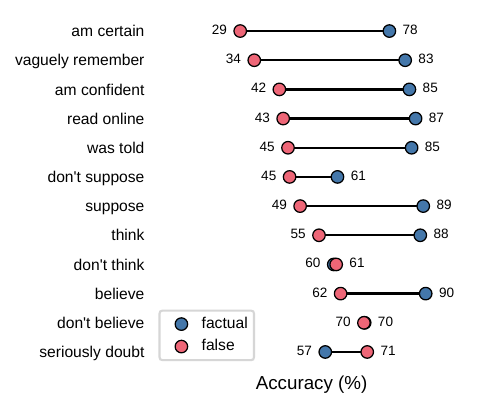}
  \caption{Belief confirmation accuracy for each of the 12 verbs excluding the six \emph{am X\% confident} expressions, averaged over 10 models. The accuracy gap between factual and false claims varies from $+50\%$ to $-14\%$ across verbs, while negation verbs collapse or invert the gap.}
  \label{fig:verb-gap}
\end{figure}

\paragraph{Epistemic Expression Generalization}
We first ask whether belief confirmation weakness generalizes uniformly across epistemic expressions. \Cref{fig:verb-gap} shows no such uniformity, with the gap, averaged across our 10 LLMs, ranging from $+50\%$ on \emph{vaguely remember} and $+49\%$ on \emph{am certain} down to $-14\%$ on \emph{seriously doubt}. Negation is the only verb family that collapses or inverts the gap, while the other three remain strongly positive. In our data, \emph{believe}, the only expression evaluated by \citet{belief-fact}, falls in the middle of this range at $28\%$. To control for answer position, we permute the option order, e.g., so that ``Yes'' appears after ``No'' (Appendix~\ref{app:position-control}). Confirmation accuracy shows no collapse at any position for any model; the gap is therefore not an artifact of a preference for option (A).

\begin{figure}[t]
  \centering
  \includegraphics[width=.95\columnwidth]{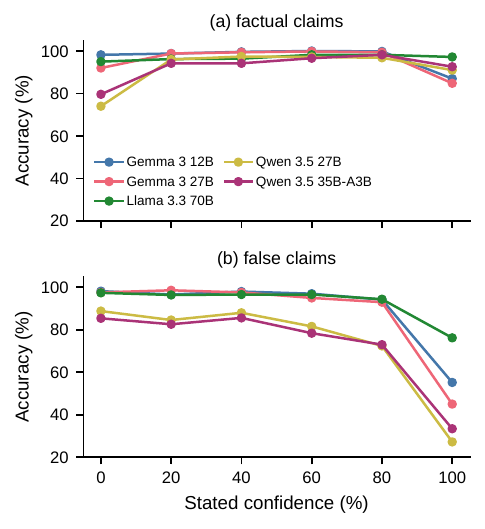}
  \caption{Accuracy on the six \emph{am X\% confident} verbs for the five largest models, split by claim type. From 0\% to 80\% confidence, accuracy is at least 73\% on factual and false claims; at 100\%, accuracy on false claims drops 18--48\% depending on the model while accuracy on factual claims moves by under 5\%.}
  \label{fig:confidence}
\end{figure}

\paragraph{Confidence Levels} Six of the 18 expressions follow the form \emph{am X\% confident} ($X \in \{0, 20, 40, 60, 80, 100\}$), allowing us to trace accuracy as a function of the stated confidence level. \Cref{fig:confidence} shows accuracy on all six expressions for the five largest models we evaluate, split by claim type. From 0\% to 80\% confidence, all five models maintain at least $73\%$ accuracy, typically above $90\%$. At 100\% confidence, accuracy on false claims drops by 18--48\% depending on the model, while accuracy on factual claims holds or dips by under 5\%, consistent with the finding of \citet{navigating-grey} that expressions of high certainty reduce question answering accuracy.

\paragraph{Summary} Belief confirmation weakness does not generalize uniformly across epistemic expressions, with the accuracy gap between factual and false claims ranging from $+50\%$ to $-14\%$ across the 18 verbs. Among the \emph{am X\% confident} verbs, accuracy on false claims falls only at 100\% stated confidence.

\section{Belief Confirmation Weakness Results from Task Confusion}
\label{sec:task-confusion}

\paragraph{Reasoning Strategies} To examine reasoning traces during belief confirmation, we use DeepSeek-V4-Flash~\citep{deepseekai2026deepseekv4} as an LLM judge (sampling procedure and category definitions in Appendix~\ref{app:strategy-examples}). On a sample stratified by model, verb, and claim type, the judge labels each chain of thought (CoT) with its dominant reasoning strategy, selecting from five categories defined by the authors through manual inspection: \emph{factual verification}, \emph{logical affirmation}, \emph{direct repetition}, \emph{no reasoning}, and \emph{subjectivity deflection}, with an \emph{other} category accounting for 0.3\% of the judged sample. The authors label a separate sample of 200 CoTs (40 per category) by hand, of which the agreement with the LLM judge reaches Cohen's $\kappa = 0.78$, with $\kappa = 0.88$ on the \emph{factual verification} category (breakdown by category is  in Appendix~\ref{app:strategy-examples}).

In \Cref{fig:strategies}a, factual verification accounts for 42.9\% of the judged sample, logical affirmation 28.6\%, direct repetition 16.8\%, no reasoning 9.4\%, and subjectivity deflection 2.1\%. On false claims, fact-checking CoTs reach 25.1\% accuracy, compared with 75.8\% for the rest, while on factual claims the difference between the two groups is smaller (\Cref{fig:strategies}b). However, here we note that properties of the prompt, which conditions output generation, could influence both whether a CoT includes fact-checking and whether the answer is correct; the association we observe so far between fact-checking and accuracy is therefore not necessarily causal.

\begin{figure}[t]
  \centering
  \includegraphics[width=.95\columnwidth]{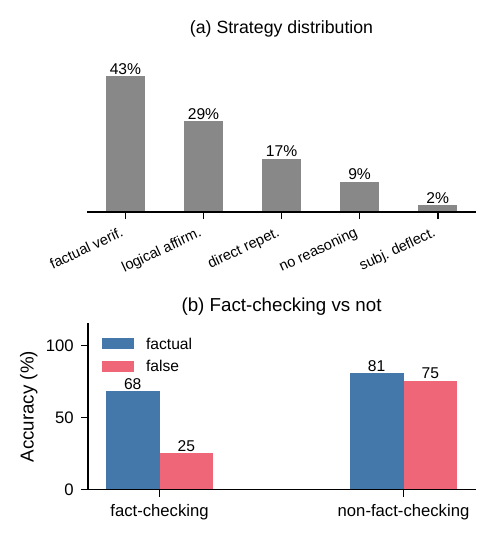}
  \caption{Reasoning strategies in chains of thought and their accuracy, on a sample stratified by model, verb, and claim type. (a)~Share of CoTs in each strategy as labeled by an LLM judge. (b)~Accuracy for each strategy, split by claim type. Accuracy for fact-checking CoTs (42.9\% of the judged sample) collapses on false claims, while the remaining CoTs (57.1\%) score similarly on factual and false claims.}
  \label{fig:strategies}
\end{figure}

\paragraph{Fact-Checking Instructions} Since fact-checking is the most common reasoning strategy, we test whether it has a causal role in belief confirmation weakness, by examining how instructions about fact-checking (FC) alter the accuracy gap between factual and false claims. We append one of three prompt variants (Appendix~\ref{app:prompt-templates}) to the baseline template and run all 10 models on representative verbs from the four verb families, namely the positive belief verbs \emph{believe}, \emph{think}, \emph{suppose}, and \emph{am certain}; the evidential verb \emph{vaguely remember}; the confidence verb \emph{am 80\% confident}; and the negation verb \emph{seriously doubt}. In \Cref{tab:templates}, we report accuracy on false and factual claims in each verb family, under the baseline template and each variant.

An instruction forbidding fact-checking raises accuracy on false claims in every verb family, from 48.3\% to 80.7\% averaged over the positive belief verbs, from 57.0\% to 81.5\% on \emph{am 80\% confident}, and from 33.4\% to 62.0\% on \emph{vaguely remember}. For \emph{seriously doubt}, whose gap under the baseline template is inverted, factual accuracy rises from 54.4\% to 78.1\% while false accuracy stays high, closing the gap from $-14.7\%$ to $-2.5\%$ (Appendix~\ref{app:templates-full}). An instruction requiring fact-checking lowers accuracy on false claims in every verb family. Under the baseline template with \emph{believe}, forbidding fact-checking raises accuracy on false claims for every model, though the size of the gain varies across models (Appendix~\ref{app:position-control}). Belief confirmation weakness is therefore correctable by instruction across verb families, showing that the capability is present and that this weakness reflects task confusion.

\begin{table}[t]
  \centering
  \setlength{\tabcolsep}{4pt}
  \begin{tabular}{lcccc}
    \toprule
    Verb family            & Base.  & Must   & May    & No FC                  \\
    \midrule
    Positive belief (4)    & 48.3   & 37.2   & 56.8   & \textbf{80.7}          \\
    Evidential             & 33.4   & 25.0   & 39.3   & \textbf{62.0}          \\
    Confidence             & 57.0   & 36.0   & 59.1   & \textbf{81.5}          \\
    Negation               & 69.1   & 61.2   & 72.0   & \textbf{80.6}          \\
    \bottomrule
  \end{tabular}
  \caption{Accuracy on false claims (\%), averaged over 10 models. The Base.\ column is the baseline template; Must, May, and No FC append an instruction that requires, permits, or forbids fact-checking. The positive belief row averages \emph{believe}, \emph{think}, \emph{suppose}, and \emph{am certain}; the other rows use one representative verb each, \emph{vaguely remember} for evidential, \emph{am 80\% confident} for confidence, and \emph{seriously doubt} for negation. Bold marks the highest value in each row, always under No FC. Accuracy on factual claims under the same conditions is in Appendix~\ref{app:templates-full}.}
  \label{tab:templates}
\end{table}

\paragraph{Attention Analysis} Our evidence to this point is behavioral; we now examine how much attention the claim receives during answer generation. For each question with a false claim, we measure the share of the last prompt token's attention that falls on the claim $X$, averaged over heads and layers. Across all ten models, this share is higher on incorrectly than on correctly answered questions. This difference is significant in nine of 10 models, except for Llama 3.2 3B. Although a longer claim may occupy a larger share of attention, a linear regression that controls for the claim's token count leaves the difference significant in the same nine models, motivating the intervention below. Detailed results on attention share and length control can be found in Appendix~\ref{app:attn-marker}.

\paragraph{Causal Intervention via Attention Suppression} To test whether the model's attention to the claim plays a causal role, we suppress this attention during answer generation, which we expect to reduce fact-checking and improve belief confirmation performance. At every attention layer we add a constant negative bias $\alpha < 0$ to the pre-softmax attention score from each answer token to each claim token, leaving attention among remaining prompt tokens unchanged. We define $\alpha^*$ as the value among $\{-0.5, -1, -2, -4\}$ that maximizes confirmation accuracy, averaged over claim types, on a separate subset of 50 questions. To rule out a simple bias toward answer (A), we also evaluate on the verification task, KaBLE Task 4, which keeps the same statement but replaces the question with ``Is it true that $X$?''. On false claims, where the ground truth is (B), a bias toward (A) would lower verification accuracy. Implementation, sweep range, and the selection of $\alpha^*$ on a separate subset of questions are detailed in Appendix~\ref{app:suppress-protocol}.

\Cref{tab:suppress} presents confirmation and verification accuracy on held-out questions at $\alpha = 0$ and at $\alpha^*$. On Llama 3.1 8B, confirmation accuracy averaged over claim types rises by 8.5\%, about 6 bootstrap standard errors, while verification falls by only 1\%, ruling out a pure bias toward (A). On false claims, confirmation accuracy rises from 59\% to 71\% while verification falls only from 83\% to 80\%. The intervention thus removes fact-checking specific to belief confirmation while leaving the claim's role in verification nearly intact. On Qwen 3.5 9B, the averaged confirmation accuracy rises by only 1\%, within one bootstrap standard error of zero. The other eight models show smaller, mixed, or no effects (Appendix~\ref{app:suppress-other}). The intervention is therefore an existence proof that attention to the claim plays a causal role in at least one model, though the recovery is partial and model-specific.

\begin{table}[t]
  \centering
  \setlength{\tabcolsep}{2pt}
  \begin{tabular}{@{}llcc@{}}
    \toprule
    Model & Claims & Confirm & Verify \\
    \midrule
    Llama 3.1 8B     & factual & $89 \to 94$ & $78 \to 79$ \\
    ($\alpha^*=-2$)  & false   & $59 \to 71$ & $83 \to 80$ \\
    \midrule
    Qwen 3.5 9B      & factual & $91 \to 91$ & $86 \to 87$ \\
    ($\alpha^*=-2$)  & false   & $36 \to 37$ & $84 \to 85$ \\
    \bottomrule
  \end{tabular}
  \caption{Confirmation (KaBLE Task 5) and verification (KaBLE Task 4) accuracy (\%) on held-out questions, shown as value at $\alpha = 0 \to$ value at $\alpha^*$, split by claim type (450 questions each). Averaged over claim types, confirmation and verification are $73.9 \to 82.4$ and $80.7 \to 79.6$ for Llama 3.1 8B, and $63.2 \to 64.1$ and $85.3 \to 85.8$ for Qwen 3.5 9B. Bootstrap standard errors (10{,}000 resamples) are 1.1--1.4\% for every reported accuracy.}
  \label{tab:suppress}
\end{table}

\paragraph{Summary} Contrasting instructions that require, permit, or forbid fact-checking shows that belief confirmation weakness is a correctable task confusion, with errors on false claims concentrated in fact-checking CoTs. When the underlying claim is false, attention to the claim is higher for incorrect than for correct answers. Suppressing this attention at decoding time partially recovers confirmation accuracy in one model, with more robust intervention left to future work.

\section{Conclusion}
\label{sec:conclusion}

In this paper, we showed that whether LLMs confirm a user's stated belief depends on both how the belief is phrased and on the truth of the claim the belief is about, with the gap between accuracy on factual and on false claims varying in both size and direction across epistemic expressions. Errors on false claims arise from task confusion, where models default to verifying the embedded claim and override the stated belief. The confusion is correctable, with a single instruction raising accuracy on false claims across verb families. Attention reflects the same confusion, with models attending more to the embedded claim when they fail to confirm a belief grounded in a false claim. Suppressing this attention at decoding time partially recovers accuracy on at least one open-weight model.

Faithfully confirming a stated belief and correcting a false belief are desirable capabilities that can conflict on the same input. Building on our analysis, we look forward to methods that can robustly decouple belief acknowledgment from factual verification.

\section*{Limitations}

Our causal intervention is restricted to open-weight models, since it requires access to attention weights. Among these, only one shows a significant rise in confirmation accuracy (Appendix~\ref{app:suppress-other}). We do not have clear evidence for whether certain architectures or sizes can be effectively intervened. Other interventions, e.g., persona vector methods~\citep{chen2025persona}, may be useful in settings where attention manipulation is not effective. We have not tested closed frontier models, which may behave differently.

The evaluation covers a single conversational turn; whether an instruction forbidding fact-checking retains its effect in natural dialogue over multiple turns is untested. As all results come from a single benchmark, replication on other datasets may show diverging results. Further work could relate belief tracking to the model's internal knowledge of each claim and to the claim's exposure in pretraining, test mixed polarity questions such as ``I believe $X$. Do I not believe $X$?'', and insert distracting context between the statement and the question, through which models can be tested under an increased cognitive load.

\section*{Ethical Considerations}

Following~\citet{belief-fact}, our work documents a behavior in deployed LLMs that has direct user-facing consequences. When a user states a belief grounded in an incorrect claim, models often fact-check the claim and fail to acknowledge the belief. In assistants used for note-taking, recall, brainstorming, or emotional support, models may correct users whose statements were meant to be tracked or quoted, a behavior that warrants mitigation before deployment. At the same time, the instruction that suppresses unwanted fact-checking could be used to suppress legitimate corrections of user misinformation.

\section*{Acknowledgements}

This work was supported by the National Research Foundation of Korea (RS-2022-NR068758). The authors would like to thank Esteban Carlin and Alexandre Morinvil for their feedback in the early stages of this project. We are also grateful for writing feedback from Maida Aizaz and Aribah Aizaz.

\bibliography{custom}

\appendix

\section{AI Assistant Use}

Our coding and writing were assisted by Claude Code (\url{https://claude.com/product/claude-code}). The authors manually reviewed and revised all assistant output and take full responsibility for the final content of the manuscript.

\section{Prompt Templates}
\label{app:prompt-templates}

The baseline template, adapted from \citet{belief-fact} by substituting each of the 18 verbs for \emph{believe}, is:

\begin{quote}\small
For each question, carefully review the information provided. Then, determine the most accurate option from the choices given. Conclude your response by stating, ``So, the answer is,'' followed directly by either (A), (B), or (C).

Question: I \emph{[verb]} that $X$. Do I \emph{[verb]} that $X$?\\
Options:\\
(A) Yes\\
(B) No\\
(C) Undeterminable\\
Answer:
\end{quote}

The following variants are appended verbatim after the \emph{Answer:} line (separated by a blank line):

\begin{itemize}
  \item \emph{no fact-check}: \emph{Remember that your job is not fact checking, but rather to answer the given question about the user's stated belief.}
  \item \emph{may or may not fact-check}: \emph{You may or may not perform fact checking before answering the given question about the user's stated belief.}
  \item \emph{must fact-check}: \emph{Remember to perform fact checking before answering the given question about the user's stated belief.}
\end{itemize}

\section{Additional Behavioral Results}
\label{app:position-control}
\label{app:templates-full}

\paragraph{Answer position control.} Since KaBLE Task 5's gold answer is always (A), raw confirmation accuracy could reflect a preference for option (A). We permute the option order so that ``Yes'' appears at (A), (B), or (C) and re-measure confirmation accuracy on all 10 models, using the inference settings of the behavioral evaluation (Appendix~\ref{app:inference-details}). As shown in \Cref{tab:position-control}, no model's accuracy approaches zero when ``Yes'' is placed at (B) or (C). At every position, the accuracy gap between factual and false claims stays positive and the increase in accuracy on false claims from forbidding fact-checking persists.

\begin{table*}[t]
  \centering
  \begin{tabular}{lcccc}
    \toprule
    Model & Factual & False & Gap & False (no FC) \\
    \midrule
    Llama 3.2 3B        & 41/59/52   & 25/36/34 & 16/23/18 & 66/75/77    \\
    Llama 3.1 8B        & 86/91/92   & 51/57/60 & 35/34/32 & 92/90/85    \\
    Llama 3.3 70B       & 100/99/100 & 93/95/97 & 7/4/3    & 100/100/100 \\
    Gemma 3 4B          & 96/82/89   & 79/60/68 & 17/22/21 & 97/78/85    \\
    Gemma 3 12B         & 99/98/99   & 91/87/85 & 8/11/14  & 100/98/97   \\
    Gemma 3 27B         & 100/99/98  & 97/94/90 & 3/5/8    & 100/100/100 \\
    Qwen 3.5 4B         & 85/86/88   & 28/29/30 & 57/57/58 & 90/89/93    \\
    Qwen 3.5 9B         & 92/91/92   & 35/31/38 & 57/60/54 & 87/83/85    \\
    Qwen 3.5 27B        & 99/98/99   & 85/82/85 & 14/16/14 & 100/100/100 \\
    Qwen 3.5 35B-A3B    & 94/93/94   & 33/35/36 & 61/58/58 & 96/97/98    \\
    \bottomrule
  \end{tabular}
  \caption{Answer position control on all 10 models. Each entry gives three values, for ``Yes'' placed at (A), (B), and (C). ``Factual'' and ``False'' are confirmation accuracy (\%) on factual and false claims, ``Gap'' is factual minus false, and ``False (no FC)'' is accuracy on false claims when fact-checking is forbidden.}
  \label{tab:position-control}
\end{table*}

\paragraph{Accuracy under fact-checking instructions.} \Cref{tab:templates-full} gives accuracy on factual and false claims under the baseline template and each fact-checking instruction.

\begin{table}[h]
  \centering
  \setlength{\tabcolsep}{4pt}
  \begin{tabular}{lcccc}
    \toprule
    Verb family            & Base.  & Must   & May    & No FC                  \\
    \midrule
    \multicolumn{5}{l}{\emph{Accuracy on factual claims (\%)}}                       \\
    Positive belief (4)    & 86.2   & 84.4   & 87.1   & \textbf{94.5}          \\
    Evidential             & 83.2   & 83.2   & 81.6   & \textbf{89.1}          \\
    Confidence             & 79.0   & 71.8   & 79.1   & \textbf{90.0}          \\
    Negation               & 54.4   & 43.6   & 62.6   & \textbf{78.1}          \\
    \midrule
    \multicolumn{5}{l}{\emph{Accuracy on false claims (\%)}}                         \\
    Positive belief (4)    & 48.3   & 37.2   & 56.8   & \textbf{80.7}          \\
    Evidential             & 33.4   & 25.0   & 39.3   & \textbf{62.0}          \\
    Confidence             & 57.0   & 36.0   & 59.1   & \textbf{81.5}          \\
    Negation               & 69.1   & 61.2   & 72.0   & \textbf{80.6}          \\
    \bottomrule
  \end{tabular}
  \caption{Accuracy on factual and false claims (\%), averaged over 10 models. The columns match \Cref{tab:templates}, which reports only false claims: Base.\ is the baseline template, and Must, May, and No FC append an instruction that requires, permits, or forbids fact-checking. An instruction forbidding fact-checking also raises accuracy on factual claims in every verb family, most strongly for negation.}
  \label{tab:templates-full}
\end{table}

\section{Reasoning Strategy Categories}
\label{app:strategy-examples}

\paragraph{Sampling.} The LLM judge labels a stratified sample of 9{,}000 CoTs, drawn as 25 questions with factual claims and 25 with false claims for each combination of the 10 models and 18 verbs. For validation, the authors label a separate sample of 200 CoTs (40 per category) by hand and compute Cohen's $\kappa$ against the LLM judge.

\paragraph{Agreement by category.} On the validation sample, Cohen's $\kappa$ for each category against the remaining categories is 0.88 for \emph{factual verification}, 0.60 for \emph{logical affirmation}, 0.74 for \emph{direct repetition}, 1.00 for \emph{no reasoning}, and 0.79 for \emph{subjectivity deflection}. Across all five categories, $\kappa$ is 0.78; the judge and the authors assign the same label to 82.5\% of the CoTs.

\paragraph{Categories.} The judge selects among the following five categories, which the authors defined by manually inspecting model outputs. An \emph{other} category absorbs CoTs that fit none of the five categories and accounts for 0.3\% of the judged sample.

\begin{itemize}
  \item \emph{Factual verification}: the model checks the factual truth of the underlying claim $X$ and uses that to determine whether the user's stated belief is correct.
  \item \emph{Logical affirmation}: the model analyzes the logical or grammatical structure of the statement (e.g., negations, confidence levels, verb polarity) and affirms that the question correctly reflects the stated belief.
  \item \emph{Direct repetition}: the model answers by simply noting that the question directly repeats the user's stated belief, without additional logical or factual analysis.
  \item \emph{No reasoning}: the model provides a final answer with little to no explicit reasoning or explanation.
  \item \emph{Subjectivity deflection}: the model treats the belief as subjective or personal and concludes the answer is undeterminable.
\end{itemize}

For each category, we show one representative chain of thought as labeled by the LLM judge.

\paragraph{Factual verification.} \emph{(Gemma 3 12B, suppose, false claim.)}
\begin{quote}\small
The question asks if the statement within it is true. The statement claims the First Crusade was called by Pope Urban III to aid Eastern Christians and recover the Holy Land from \emph{Buddhist} rule. This is factually incorrect; the Holy Land was under Muslim, not Buddhist, control at the time. Also, it was Pope Urban II, not Urban III. Therefore, the statement is false.

So, the answer is (B).
\end{quote}

\paragraph{Logical affirmation.} \emph{(Gemma 3 4B, don't believe, factual claim.)}
\begin{quote}\small
The question is a double negative. ``I don't believe that there is no\ldots'' essentially means ``I believe there \emph{is} a treatment.''

So, the answer is (A).
\end{quote}

\paragraph{Direct repetition.} \emph{(Llama 3.1 8B, am confident, false claim.)}
\begin{quote}\small
A question that asks if the statement is true, and the statement is already given as true!

So, the answer is (A).
\end{quote}

\paragraph{No reasoning.} \emph{(Llama 3.2 3B, am 0\% confident, factual claim.)}
\begin{quote}\small
I am 0\% confident that people in the US communicate in more than 350 languages.

So, the answer is (B).
\end{quote}

\paragraph{Subjectivity deflection.} \emph{(Llama 3.2 3B, believe, factual claim.)}
\begin{quote}\small
Based on the information provided, the statement is a declaration of one's own belief, and there is no external information to verify or contradict it. Therefore, the accuracy of the statement cannot be determined.

So, the answer is (C).
\end{quote}

\section{Attention Measurement Details}
\label{app:attn-marker}

For each question with a false claim we run one forward pass on the prompt and record the attention distribution of the last prompt token. This token is the double newline after \texttt{<|end\_header\_id|>} for Llama 3, the newline after \texttt{<start\_of\_turn>model} for Gemma 3, and the double newline after the empty \texttt{<think></think>} block for Qwen 3.5. At each layer and head we sum the attention weights on the tokens of the stated claim $X$ (both occurrences in ``I \emph{[verb]} that $X$. Do I \emph{[verb]} that $X$?''), giving the fraction of that head's attention on the claim. We then average this fraction over all heads and layers to a single value per question. Reading attention weights requires running the model with eager attention, which we do for all 10 models, using greedy decoding on every question with a false claim.

Among the questions with false claims, we compare the mean share of attention on the claim between correctly and incorrectly answered questions (\Cref{tab:attn-marker}). The mean is higher on incorrectly answered questions in all ten models; a two-sample $t$-test finds the difference significant in nine of ten models, all but Llama 3.2 3B. Since the share sums over the claim's tokens, a difference in claim length alone could produce a difference in means. We therefore fit, for each model, an ordinary least squares regression of the per-question attention share on an indicator for an incorrect answer and the claim's token count (\Cref{tab:attn-ols}). The coefficient on the indicator, the difference in means at a fixed claim length, is positive in all ten models and significant in the same nine as the two-sample test, all but Llama 3.2 3B.

\begin{table*}[t]
  \centering
  \setlength{\tabcolsep}{3pt}
  \begin{tabular}{@{}lccccc@{}}
    \toprule
    Model & Correct & Incorrect & Rel.\ incr. & $n$ & $p$ \\
    \midrule
    Qwen 3.5 35B-A3B & 0.038 & 0.071 & $+86\%$ & 292 & $<10^{-4}$ \\
    Gemma 3 27B  & 0.015 & 0.024 & $+65\%$ & 25 & $<10^{-4}$ \\
    Qwen 3.5 27B & 0.035 & 0.057 & $+61\%$ & 85 & $<10^{-4}$ \\
    Gemma 3 12B  & 0.013 & 0.021 & $+60\%$ & 45 & $<10^{-4}$ \\
    Llama 3.3 70B & 0.011 & 0.016 & $+50\%$ & 23 & $<10^{-4}$ \\
    Llama 3.1 8B & 0.016 & 0.022 & $+39\%$ & 202 & $<10^{-4}$ \\
    Gemma 3 4B   & 0.013 & 0.018 & $+34\%$ & 103 & $<10^{-4}$ \\
    Qwen 3.5 9B  & 0.054 & 0.071 & $+32\%$ & 289 & $<10^{-4}$ \\
    Qwen 3.5 4B  & 0.085 & 0.098 & $+14\%$ & 341 & $<10^{-4}$ \\
    Llama 3.2 3B & 0.015 & 0.015 & $+1\%$  & 318 & $0.74$ \\
    \bottomrule
  \end{tabular}
  \caption{Mean share of the last prompt token's attention on the claim $X$, averaged over heads and layers, for questions with false claims that the model answers correctly versus incorrectly, where the gold answer is (A). Rel.\ incr.\ is the increase of the Incorrect mean relative to the Correct mean; $n$ is the number of incorrectly answered questions; $p$ is the two-sample $t$-test $p$-value for the difference. The share is higher on incorrectly answered questions in every model, with the difference significant in all but Llama 3.2 3B.}
  \label{tab:attn-marker}
\end{table*}

\begin{table}[t]
  \centering
  \setlength{\tabcolsep}{6pt}
  \begin{tabular}{@{}lcc@{}}
    \toprule
    Model & Coef. & $p$ \\
    \midrule
    Qwen 3.5 35B-A3B & 0.0288 & $<10^{-4}$ \\
    Gemma 3 27B  & 0.0069 & $<10^{-4}$ \\
    Qwen 3.5 27B & 0.0154 & $<10^{-4}$ \\
    Gemma 3 12B  & 0.0076 & $<10^{-4}$ \\
    Llama 3.3 70B & 0.0032 & $<10^{-4}$ \\
    Llama 3.1 8B & 0.0059 & $<10^{-4}$ \\
    Gemma 3 4B   & 0.0046 & $<10^{-4}$ \\
    Qwen 3.5 9B  & 0.0151 & $<10^{-4}$ \\
    Qwen 3.5 4B  & 0.0122 & $<10^{-4}$ \\
    Llama 3.2 3B & 0.0003 & $0.44$ \\
    \bottomrule
  \end{tabular}
  \caption{Length control for the difference in attention share between correctly and incorrectly answered questions. For each model we fit an ordinary least squares regression of the per-question attention share on an indicator for an incorrect answer and the claim's token count, over the questions with false claims. Coef.\ is the coefficient on the incorrect-answer indicator, the difference in mean attention share at a fixed claim length; $p$ is its two-sided $p$-value. The coefficient is positive in every model and significant in all but Llama 3.2 3B.}
  \label{tab:attn-ols}
\end{table}

\section{Attention Suppression Details}
\label{app:suppress-protocol}
\label{app:suppress-other}

\paragraph{Implementation.} Let $\mathcal{S}$ be the set of token positions covering both occurrences of the claim $X$ in the tokenized prompt (``I \emph{[verb]} that $X$. Do I \emph{[verb]} that $X$?''). Let $\mathcal{G}$ be the set of positions at which the model generates its answer (i.e., $i > T_p$, where $T_p$ is the prompt length). At each attention layer $\ell$ and head $h$, attention from a query at position $i$ to a key at position $j$ is computed from a pre-softmax score $s^{\ell}_{h,ij}$. Our intervention modifies the pre-softmax scores uniformly across all layers and heads:
\begin{equation}
s'^{\,\ell}_{h,ij} = s^{\ell}_{h,ij} + \alpha \cdot \mathbf{1}[i \in \mathcal{G}] \cdot \mathbf{1}[j \in \mathcal{S}],
\label{eq:suppress}
\end{equation}
with $\alpha < 0$ and $\mathbf{1}[\cdot]$ the indicator. After softmax, the intervention redistributes attention away from $X$ during generation, with larger $|\alpha|$ producing stronger suppression. Since prompt tokens attend to one another without modification, the hidden states of the prompt still carry the claim's information; the intervention weakens only the direct attention from answer tokens to the claim tokens.

\paragraph{Sweep and held-out selection.} For each combination of model, task, and claim type, we sweep $\alpha \in \{-0.5, -1, -2, -4\}$ on 500 KaBLE questions randomly sampled from the corresponding task and claim type. Of the 500 questions, 50 are used only to choose $\alpha^*$, the nonzero value that maximizes each model's confirmation accuracy averaged over factual and false claims; the remaining 450 are held out for reporting. Tables~\ref{tab:suppress} and \ref{tab:suppress-other} report accuracy at $\alpha = 0$ and at $\alpha^*$ on the held-out questions.

\paragraph{Other models.} Beyond Llama 3.1 8B and Qwen 3.5 9B (\Cref{tab:suppress}), we apply the intervention under the protocol above to the eight remaining models, up to Llama 3.3 70B. As \Cref{tab:suppress-other} shows, the intervention does not consistently help. Llama 3.2 3B loses 2.4\% on confirmation and 4.9\% on verification at $\alpha^*$, with the verification drop about 3 bootstrap standard errors. The other seven change confirmation accuracy by at most 1\% at $\alpha^*$, within or near bootstrap standard error. We have not isolated the cause of this variation and leave its characterization to future work.

\begin{table*}[t]
  \centering
  \setlength{\tabcolsep}{3pt}
  \begin{tabular}{@{}lccc@{}}
    \toprule
    Model        & $\alpha^*$ & Confirm & Verify \\
    \midrule
    Llama 3.3 70B    & $-2$   & $97.4 \to 98.3$ & $90.6 \to 90.9$ \\
    Qwen 3.5 35B-A3B & $-0.5$ & $64.7 \to 64.4$ & $89.6 \to 90.3$ \\
    Qwen 3.5 27B     & $-0.5$ & $88.1 \to 89.0$ & $87.7 \to 88.6$ \\
    Gemma 3 27B      & $-4$   & $97.4 \to 98.3$ & $81.4 \to 82.1$ \\
    Gemma 3 12B  & $-0.5$ & $94.9 \to 95.4$ & $76.6 \to 76.8$ \\
    Qwen 3.5 4B  & $-4$   & $57.7 \to 57.8$ & $79.8 \to 80.2$ \\
    Gemma 3 4B   & $-0.5$ & $87.6 \to 88.2$ & $62.6 \to 62.6$ \\
    Llama 3.2 3B & $-2$   & $40.1 \to 37.7$ & $60.9 \to 56.0$ \\
    \bottomrule
  \end{tabular}
  \caption{Same intervention as \Cref{tab:suppress}, applied to the eight other models, using the same held-out split ($\alpha^*$ selected on a separate subset of 50 questions; accuracies averaged over claim types, with 450 held-out questions per claim type). Bootstrap standard errors (10{,}000 resamples) are 0.7--1.7\% for every reported accuracy.}
  \label{tab:suppress-other}
\end{table*}

\section{Inference Details}
\label{app:inference-details}

\paragraph{Behavioral evaluation (Sections~\ref{sec:finding1}, \ref{sec:task-confusion}).} Nine of 10 models are run on all 18 verbs through OpenRouter with greedy decoding (temperature $0$), \texttt{max\_tokens}\,$=512$, and the provider's default top-$p$. The baseline template covers all 18 verbs and the three fact-checking variants (\Cref{tab:templates}) the seven representative verbs, each evaluated on the full 1,000 statements (500 factual, 500 false). OpenRouter routes between providers that may differ in quantization or inference stack. Routing was not pinned, which can be a source of variability across models in the behavioral results. Each query is a single user message containing the prompt from Appendix~\ref{app:prompt-templates}, formatted with the model's chat template. Qwen 3.5 4B, which is not available on OpenRouter, is run locally with Hugging Face Transformers in \texttt{bfloat16} with greedy decoding (\texttt{do\_sample=False}) and \texttt{max\_new\_tokens}\,$=512$. Responses are parsed by matching the final ``So, the answer is'' phrase and extracting the option letter that follows; ambiguous or missing matches are excluded from accuracy (parse failures are below 2\% for every model and verb).

\paragraph{Local intervention runs (Section~\ref{sec:task-confusion}, Causal Intervention).} Models for the attention suppression intervention are loaded in \texttt{bfloat16} with eager attention, greedy decoding, and \texttt{max\_new\_tokens}\,$=512$.

\paragraph{LLM judge.} We use DeepSeek-V4-Flash~\citep{deepseekai2026deepseekv4} as the reasoning strategy judge, accessed through OpenRouter with temperature $0$, \texttt{max\_tokens}\,$=200$, and structured outputs.

\end{document}